\documentclass{article} %
\usepackage{subfigure} 
\usepackage{nips13submit_e,times}
\usepackage{hyperref}
\usepackage{url}

\usepackage[pdftex]{graphicx}
\usepackage{natbib}
\setcitestyle{numbers,square}
\usepackage{algorithm}
\usepackage{algorithmic}
\usepackage{fancyvrb}
\usepackage{verbatim}
\usepackage{wrapfig}
\usepackage{amsmath}
\usepackage{color}
\usepackage{stfloats}
\graphicspath{{plots/}}

\newcommand{\name}{Compressed Vector Machine}

\newcommand{\argmax}{\operatornamewithlimits{argmax}}

\newcommand{\x}{\mathbf{x}}

\newcommand{\w}{\mathbf{w}}

\newcommand{\ab}{\boldsymbol{\alpha}}
\newcommand{\bl}{\boldsymbol{\beta}}

\newcommand{\y}{\mathbf{y}}

\newcommand{\Ib}{\mathbf{I}}
\newcommand{\Kb}{\mathbf{K}}

\newcommand{\Sb}{\mathbf{S}}

\newcommand{\Db}{\mathbf{D}}

\title{Compressed Support Vector Machines}

\author{
Zhixiang (Eddie) Xu \\
\texttt{xuzx@cse.wustl.edu} \\
\And
Jacob R. Gardner \\
\texttt{gardner.jake@gmail.com} \\
\AND
Stephen Tyree \\
\texttt{swtyree@wustl.edu} \\
\And
Kilian Q. Weinberger \\
\texttt{kilian@wustl.edu} \\
\AND
\textnormal{Department of Computer Science \& Engineering} \\
Washington University in St. Louis \\
St. Louis, MO, USA \\
}

\nipsfinalcopy %

\begin{document}

\maketitle

\begin{abstract}
Support vector machines (SVM) can classify data sets along highly non-linear decision boundaries because of the kernel-trick. This expressiveness comes at a price: During test-time, the SVM classifier needs to compute the kernel inner-product between a test sample and all support vectors. With large training data sets, the time required for this computation can be substantial.
In this paper, we introduce a post-processing algorithm, which \emph{compresses} the learned SVM model by reducing and optimizing support vectors. We evaluate our algorithm on several medium-scaled real-world data sets, demonstrating that it maintains high test accuracy while reducing the test-time evaluation cost by several orders of magnitude---in some cases from hours to seconds. 
It is fair to say that most of the work in this paper was previously been invented by Burges and Sch\"olkopf almost 20 years ago. For most of the time during which we conducted this research, we were unaware of this prior work. However, in the past two decades, computing power has increased drastically, and we can therefore provide empirical insights that were not possible in their original paper. 

\end{abstract}

\section{Introductions}
Support Vector Machines (SVM) are arguably one of the great success stories of machine learning and have been used in many real world applications, including email spam classification~\cite{drucker1999support}, face recognition~\cite{heisele2001face} and gene selection~\cite{guyon2002gene}. 
In real world applications, the evaluation cost (in terms of memory and CPU) during test-time is of crucial importance. This is particularly prominent in settings with strong resource constraints (\emph{e.g.} embedded devices, cell phones or tablets)  or frequently repeated tasks (\emph{e.g.} webmail spam classification, web-search ranking, face detection in uploaded images), which can be performed billions of times per day. 
Reducing the resource requirements to classify an input can reduce hardware costs, enable product improvements, and help curb power consumption.

Test-time cost is determined mainly by two components: classifier evaluation and feature extraction cost. Reducing feature extraction cost has recently obtained a significant amount of attention~\cite{busa2012fast,chen2011,Lefakis2010,pujara2011using,Saberian2010,wang2012local,xu2013cstc,xu2013afr}. These approaches reduce the test-time cost in scenarios where features are heterogeneous, extracted on-demand, and are significantly more expensive to compute than the classifier evaluation.

In this paper, we focus on the other common scenario where the classifier evaluation cost dominates the overall test-time cost. Specifically, we focus on kernel support vector machine (SVM)~\cite{scholkopf2001learning}. Kernel computation can be expensive because it is linear in the number of support vectors and, in addition, often requires expensive exponentiation (\emph{e.g.} for the radial basis or $\chi^2$ kernels). 
Previous work has reduced the classifier complexity by selecting few support vectors through budgeted training~\cite{dekel2008forgetron,wang2012breaking} or with heuristic selection prior to learning~\cite{keerthi2006building}.

We describe an approach that does not \emph{select} support vectors from the training set, but instead \emph{learns} them to match a pre-defined SVM decision boundary. Given an existing SVM model with $m$ support vectors, it learns $r\!\ll\! m$ ``artificial support vectors'', which are not originally part of the training set. The resulting model is a standard SVM classifier (thus can be saved, for example, in a LibSVM~\cite{chang2011libsvm} compatible file). Relative to the original model, it has comparable accuracy, but it is up to several orders of magnitudes smaller and faster to evaluate. 
We refer to our algorithm as \name{} (CVM) and demonstrate on eight real-world data sets of various size and complexity that it achieves unmatched accuracy vs. test-time cost trade offs.

\section{Related Work}
\emph{Burges and Sch\"olkopf~\cite{burges1997improving} invented Compressed Vector Machines long before us. While we conducted our research, we were not aware of their work until very late during the final stages of paper writing. We still consider our perspective and additional experiments valuable and decided to post our results as a techreport. However we do want to emphasize that all academic credit should go to them who were clearly ahead of us.}

Reducing test-time cost has recently attracted much attention. Much work \cite{busa2012fast,chen2011,grubbspeedboost,Lefakis2010,pujara2011using,Saberian2010,wang2012local,xu2013cstc} focuses on scenarios where features are extracted on-demand and the extraction cost dominates the overall test-time cost. Their objective is to minimize the feature extraction cost. 

Model compression was pioneered by \citep{bucilua2006model}. Our work was inspired by their vision, however it differs substantially, as we do not focus on ensembles of classifiers and instead learn a model compressor explicitly for SVMs. More recently, \cite{xu2013afr} introduces an algorithm to reduce the test-time cost specifically for the SVM classifier. However, similar to the approaches mentioned above, they focus on learning a new representations consisting of cheap non-linear features for linear SVMs.

\citep{dekel2008forgetron} propose an algorithm to limit the memory usage for kernel based online classification. Different from our approach, their algorithm is not a post-process procedure, and instead they modify the kernel function directly to limit the amount of memory the algorithm uses. Similar to \cite{dekel2008forgetron}, \citep{wang2012breaking} also focusses on online kernel SVM, and attacks primarily the training time complexity.  

Of particular relevance is \cite{keerthi2006building}, which, specifically reduces the SVM evaluation cost by reducing the number of support vectors.  Heuristics are used to select a small subset of support vectors, up to a given budget, during training time, thus solving an approximate SVM optimization. In contrast, our method is a post-processing compression to the regular SVM.  We begin from an exact SVM solution and compress the set of support vectors by choosing and optimizing over a small set of support vectors to approximate the optimal decision boundary.  This post-processing optimization framework renders unmatched accuracy and cost performance. Similar approaches have successfully learned pseudo-inputs for compressed nearest neighbor classification sets \citep{kusner2014stochastic} and sparse Gaussian process regression models \citep{snelson2005sparse}.

\section{Background}
Let the data consist of input vectors $\{\x_1,\dots,\x_n\} \in {\cal R}^d$ and corresponding labels $\{y_1,\dots, y_n\} \in \{-1, +1\}$. For simplicity we assume binary classification in the following section, but our algorithm is easily extended to multi-class settings using one-vs-one~\cite{shawe2000support}, one-vs-all~\cite{rifkin2004defense}, or DAG~\cite{platt2000large} approaches, and results are included for several multi-class datasets.

\textbf{Kernel support vector machines.}
SVMs are popular for their large margin enforcement, which leads to good generalization to unseen test data, and their formulation as a convex quadratic optimization problem, guaranteeing a globally optimal solution. Most importantly, the \emph{kernel-trick}~\cite{scholkopf2001learning} may be employed to learn highly non-linear decision boundaries for data sets that are not linearly separable. Specifically, the kernel-trick maps the original feature space $\x_i$ into a higher (possibly infinite) dimensional space $\phi(\x_i)$.

SVMs learn a hyperplane in this higher dimensional space by maximizing the margin $\frac{1}{\|\w\|}$ and penalizing training instances on the wrong side of the hyperplane,
\begin{align}
	\min_{\w,b} \|\w\| + C\sum_i^n \Big( \max\big(1 - y_i \w^\top \phi(\x_i)+b, 0\big) \Big)^2, \label{eq:svmprimal}
\end{align}
where $b$ is the bias, and $C$ trades-off regularization/margin and training accuracy. Note that we use the quadratic hinge loss penalty and thus~(\ref{eq:svmprimal}) is differentiable. The power of the kernel trick is that the higher dimensional space $\phi(\x_i)$ never needs to be expressed explicitly, because~(\ref{eq:svmprimal}) can be formulated in terms of inner products between input vectors. Let a matrix $\Kb$ denote these inner products, where $\Kb_{ij} = \phi(\x_i)^\top \phi(\x_j)$, and $\Kb$ is the training \emph{kernel matrix}. The optimization in~(\ref{eq:svmprimal}) can be then expressed in terms of kernel matrix $\Kb$ in the dual form:
\begin{align}
	\max_{\alpha_1,\dots,\alpha_n} & \sum_{i=1}^n \alpha_i - \frac{1}{2} \sum_{i,j=1}^n \alpha_i \alpha_j y_i y_j \Kb_{ij}, 
	\;\;\;\textrm{  s.t. } \sum_{i=1}^n \alpha_i y_i = 0 \textrm{ and } \alpha_i \ge 0, \label{eq:svmdual}
\end{align} 
where $\alpha_i$ are the Lagrange multipliers.

the classification rule $f(\cdot)$ for a test input $\x_t$ can also be expressed by testing kernel $\tilde\Kb$ that consists of inner products between test inputs ${\cal E} = \{\x_t\}$ and support vectors ${\cal S} = \{\x_i|\alpha_i \neq 0\}$, $\tilde\Kb_{it} = \phi(\x_i)^\top \phi(\x_t)$, where 
\begin{align}
	f(\phi(\x_t)) = \sum_{i=1}^n \alpha_i y_i \tilde\Kb_{it} + b. \label{eq:svmpredict}
\end{align}
Note that once testing kernel $\tilde\Kb$ is computed, generating the prediction is merely a linear combination, and thus the dominating cost is computing the testing kernel itself. 

\textbf{Least angle regression.} 
LARS~\cite{efron2004least} is a widely used forward selection algorithm because of its simplicity and efficiency. Given input vectors $\x$, target labels $\y$, and the quadratic loss $\ell(\bl) = (\x\bl - \y)^2$, LARS learns to approximate targets by building up the coefficient vector $\bl$ in successive steps, starting from an all-zero vector. To minimize the loss function $\ell$, LARS initially descends on a coordinate direction that has the largest gradient,
\begin{align}
	\beta_t = \argmax_{\beta_t} \frac{\partial \ell}{\partial \beta_t}.
\end{align}
The algorithm then incorporates this coordinate into its active set. After identifying the gradient direction, LARS selects the step size very carefully. Instead of too greedy or too tiny, LARS computes a step size that a new direction \emph{outside} of the active set has the same maximum gradient as directions \emph{in} the active set. LARS then include this new direction into the active set. 

In the following iterations, LARS gradient descends on a direction that maintains the same gradient for all directions in the active set. In other words, LARS descends following an equiangular direction of all directions in the active set. The algorithm then repeats computing step-size, including new directions into the active set, and descending on an equiangular directions. This process makes LARS very efficient, as after $T$ iterations, LARS solution has exactly $T$ directions in the active set, or equivalently, only $T$ non-zero coefficients in $\bl$.

\section{Method}
In this section, we detail the CVM approach to reduce the test-time SVM evaluation cost. We regard CVM as a post-processing compression to the original SVM solution. After solving an SVM, we obtain a set of support vectors ${\cal S} = \{\x_i|\,\alpha_i \neq 0\}$, and the corresponding Lagrange multipliers $\alpha_i$. Given the original SVM solution, we can model the test-time evaluation cost explicitly. 

\textbf{Kernel SVM evaluation cost.}
Based on the prediction function~(\ref{eq:svmpredict}) we can formulate the exact SVM classifier evaluation cost.
Let $e$ denote the cost of computing a test kernel entry $\tilde\Kb_{it}$ (\emph{i.e.} kernel function of a test input $\x_t$ and a support vector $\x_i$).  We assume the computation cost is identical across all test inputs and all support vectors. As shown in~(\ref{eq:svmpredict}), generating a prediction for a testing input requires computing the kernel entry between the test input and all support vectors. The total evaluation cost is a function of the number of support vectors $n_{\textrm{sv}}$. After obtaining the kernel entries for a test point $\x_t$, prediction is simply linear combination of the kernel row $\tilde\Kb_t$ weighted by $\ab$.  The cost of computing this linear combination is very low compared to the kernel computation, and therefore the total evaluation cost $c_e = n_{\textrm{sv}}e$. We aim to reduce the size of the support vector set $n_{\textrm{sv}}$ without greatly affecting prediction accuracy.

\textbf{Removing non-support vectors.} 
Since the test-time evaluation cost is a function of the number of support vectors, the goal is to cherry-pick and optimize a subset of the optimal support vectors bounded in size by a user-specified compression ratio.
We first note that all non-support vectors can be removed during this process without affecting the full SVM solution.
If we define a design matrix $\hat{\Kb} \in {\cal R}^{n\times n}$, where $\hat{\Kb}_{ij} = y_i\Kb_{ij}$. The squared penalty SVM objective function in~(\ref{eq:svmprimal}) can be expressed 
with Lagrange parameter $\ab$ and the kernel matrix $\Kb$:
\begin{align}
	\min_{\ab,b} \Big(\max(\mathbf{1}-\hat{\Kb}\ab - \y b, 0)\Big)^2 + \ab^\top \Kb \ab. \label{eq:sqrsvmdual}
\end{align}
Since (\ref{eq:sqrsvmdual}) is a \emph{strongly} convex function, and all non-support vectors have the corresponding Lagrange multiplier $\alpha_i=0$, we can remove all non-support vectors from the optimization problem and the full SVM optimal solution stays the same.

To find an optimal subset of support vectors given the compression ratio, we re-train the SVM with only support vectors and a constraint on the number of support vectors. Note that $\ab$ are effectively the coefficients of support vectors, and we can efficiently control the number of support vectors by adding an $l_0$ norm on $\ab$. The optimization problem becomes 
\begin{align}
	\min_{\ab,b} & \Big(\mathbf{1}-\hat{\Kb}\ab - \y b\Big)^2 + \ab^\top \Kb \ab \label{eq:sqrsvmdualsv} \\
	\textrm{ s.t.} & \|\ab\|_0 \le \frac{1}{e} B_e, \nonumber
\end{align}
where $B_e$ evaluation cost budget, and consequently, $\frac{1}{e}B_e$ is the desired number of support vectors based on the budget. Note that after removing non-support vectors, we obtain a condensed matrix $\hat{\Kb} \in {\cal R}^{n_{sv} \times n_{sv}}$. 

\textbf{Forming ordinary least squares problem.}
The current form of equation (\ref{eq:sqrsvmdualsv}) can be made more amenable to optimization by rewriting the objective function as an ordinary least square problem.
Expanding the squared term, simplifying, and fixing the bias term $b$ (as it does not affect the solution dramatically), we re-format the objective function  (\ref{eq:sqrsvmdualsv}) into
\begin{equation}
	\min_{\ab} (\mathbf{1}-\y b)^\top (\mathbf{1}-\y b) - 2\ab^\top\hat{\Kb}^\top(\mathbf{1}-\y b) + \ab^\top(\hat{\Kb}^\top\hat{\Kb}+\Kb)\ab.
	\label{eq:expandsvmdual}
\end{equation}
We introduce two auxiliary variables $\Omega$ and $\bl$, where $\Omega^\top\Omega = \hat{\Kb}^\top\hat{\Kb} + \Kb$ and $\Omega^\top\bl = -\hat{\Kb}^\top (\mathbf{1}-\y b)$.
Because $\hat{\Kb}^\top\hat{\Kb}+\Kb$ is a symmetric matrix, 
we can compute its eigen-decomposition
\begin{align}
	\hat{\Kb}^\top\hat{\Kb}+\Kb=\Sb\Db\Sb^\top,
\end{align}
where $\Db$ is the diagonal matrix of eigenvalues and $\Sb$ is the orthonormal matrix of eigenvectors.
Moreover, because the matrix $\hat{\Kb}^\top\hat{\Kb}+\Kb$ is positive semi-definite, we can further decompose $\Sb\Db\Sb^\top$ into an inner product of two real matrices by taking the square root of $\Db$. Let $\Omega = \sqrt{\Db} \Sb^\top$, and we obtain a matrix $\Omega$ that satisfies $\Omega^\top\Omega = \hat{\Kb}^\top\hat{\Kb} + \Kb$. 
After computing $\Omega$, we can readily compute $\bl = -(\Omega^\top)^{-1}\hat{\Kb}^\top (\mathbf{1}-\y b)$, where $(\Omega^\top)^{-1} = \frac{1}{\sqrt{\Db}} \Sb^\top$.

With the help of the two auxiliary variables, we convert (\ref{eq:expandsvmdual}), plus a constant term%
\footnote{$(\mathbf{1}-\y b)^\top \left(\hat{\Kb} (\hat{\Kb}^\top\hat{\Kb}+\Kb)^{-1} \hat{\Kb}^\top - \Ib \right) (\mathbf{1}-\y b)$},
into least squares format.
Together with relaxation of the non-continuous $l_0$- norm constraint to an $l_1$-norm constraint, we obtain
\begin{align}
	\min_{\ab} (\Omega \ab + \bl)^2, \;\;\;
	\textrm{ s.t. } \|\ab\|_1 \le \frac{1}{e}B_e. \label{eq:readyforlars}
\end{align}

\textbf{Compressing the support vector set.}
The squared loss and $l_1$ constraint in~(\ref{eq:readyforlars}) lead naturally to the LARS algorithm.
Given a budget $B_e$, we can determine the maximum size $m$ of the compressed support vector set ($m = \frac{B_e}{e}$).
Using LARS, we start from an empty support vector set and add $m$ support vectors incrementally.
Since adding a support vector is equivalent to activating a coefficient in $\ab$ to a non-zero value, we can obtain $m$ optimal support vectors by running LARS optimization in~(\ref{eq:readyforlars}) exactly $m$ steps, where each step activates one coefficient. The resulting solution gives the optimal set of $m$ support vectors. We refer this intermediate step as \emph{LARS-SVM}. Note that this step is crucial for the problem, as this LARS-SVM solution serves as a very good initialization for the next step, which is a non-convex optimization problem.

\textbf{Gradient support vectors.}
If we interpret $\ab$ as coordinates and the corresponding columns in the kernel matrix $\Kb$ as basis vectors, then these basis vectors span an ${\cal R}^{n_{sv}}$ space in which lie predictions of the original SVM model.
In this compression algorithm, our goal is to find a lower dimensional subspace that supports good approximations of the original predictions.
After running LARS for $m$ iterations, we obtain $m$ support vectors and their coefficients $\ab$, forming an ${\cal R}^{m}$ subspace of the space spanned by the full kernel matrix.

\begin{figure}[t]
\centerline{
\includegraphics[width = .5\textwidth]{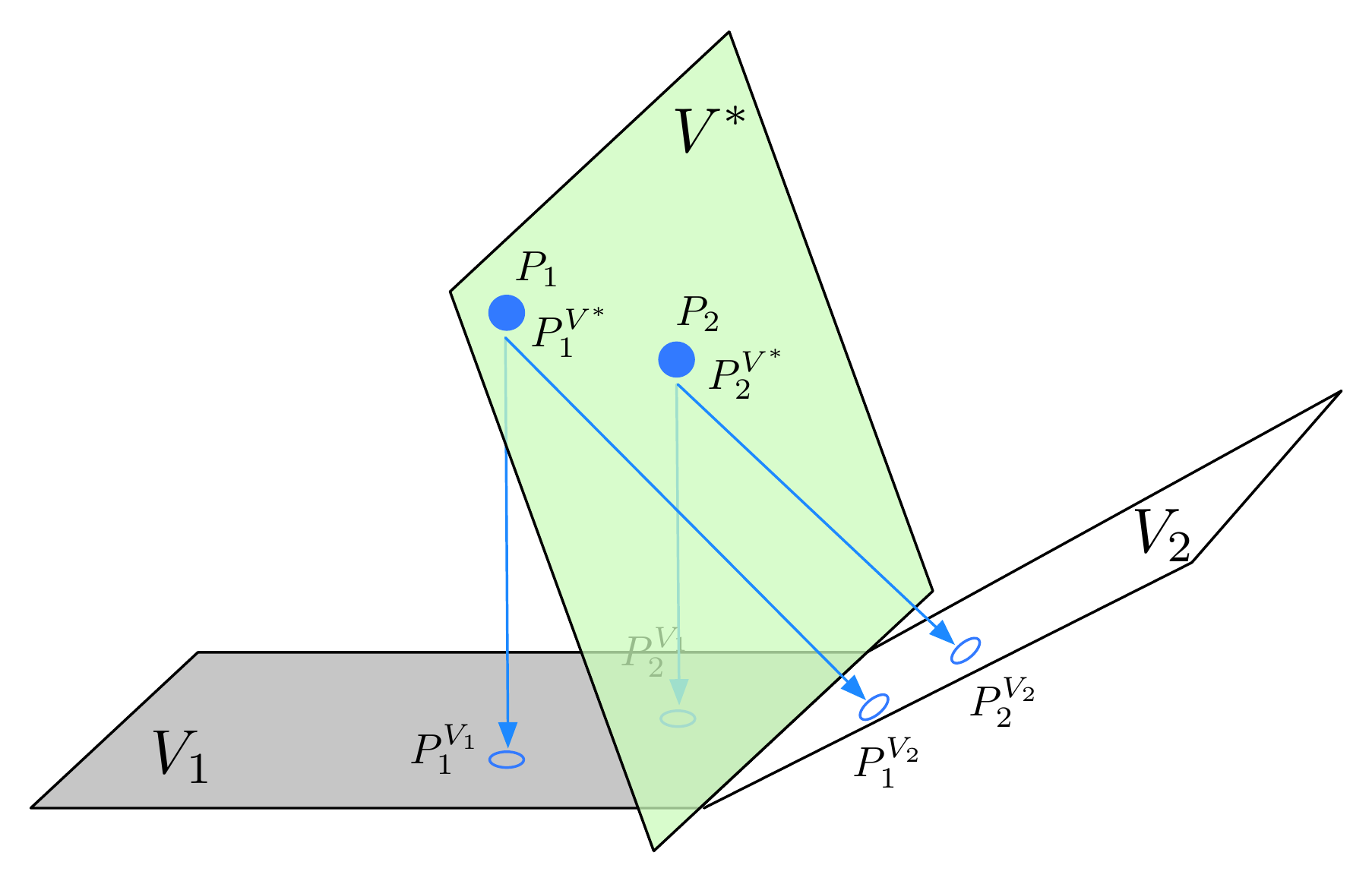}
}
\caption{Illustration of searching for a subspace $V \in {\cal R}^2$ that best approximates predictions $P_1$ and $P_2$ of training instances in ${\cal R}^3$ space. Neither $V_1$ or $V_2$, spanned by existing columns in the kernel matrix, is a good approximation. $V^*$ spanned by kernel columns computed from two \emph{artificial} support vectors is the optimal solution.
}
\label{fig:approx} 
\end{figure}

We illustrate this lower dimensional approximation in Figure~\ref{fig:approx}.
Vectors $P_1$ and $P_2$ are predictions of two training points made in the full SVM solution space (${\cal R}^3$ and spanned by three support vectors).
We want to compress the model to two support vectors by looking for a subspace $V \in {\cal R}^2$ that supports the best approximations of these two predictions.
Using existing support vectors as a basis, we can find subspaces $V_1$ and $V_2$, each spanned by a pair of support vectors.
The projections of $P_1$ and $P_2$ on plane $V_1$ ($P_1^{V_1}$ and $P_2^{V_1}$) are closest to the original predictions $P_1$ and $P_2$, and thus $V_1$ is the better approximation.
However, in this case, neither $V_1$ nor $V_2$ is a particularly good approximation.
Suppose we remove the restriction of selecting a subspace spanned by \emph{existing} basis vectors in the kernel matrix,
instead optimizing the basis vectors to yield a more suitable subspace.
In Figure~\ref{fig:approx}, this is illustrated by the optimal subspace $V^*$ which produces a better approximation to the target predictions.

Note that the basis vectors (columns of the kernel matrix) are parameterized by support vectors.
By optimizing these underlying support vectors, we can search for a better low-dimensional subspace.
If we denote $\Kb_m$ as the training kernel matrix with only $m$ columns corresponding to the support vectors chosen by LARS, and $\ab_m$ as the coefficients of these support vectors, we can formulate the search for \emph{artificial} support vector as an optimization problem. Specifically, we minimize a squared loss between approximate and full SVM predictions over all support vectors, and the parameters are support vectors.
\begin{align}
	\min_{(\x_1,\dots,\x_m)} {\cal L} = \Big(\Kb_m \alpha_m - \Kb \ab \Big)^2, 
	\label{eq:gradient}
\end{align}
where $\Kb_{ij}$ is the kernel entry, and for simplicity, we use radial basis function (RBF) kernel function ($\Kb_{ij} = e^{\frac{\|\x_i - \x_j\|^2}{2\sigma^2}}$). However, other kernel functions are equally suitable. The unconstrained optimization problem~(\ref{eq:gradient}) can be solved by conjugate gradient descent with respect to the chosen $m$ support vectors.
Since $\ab$'s are the coordinates with respect to the basis, we optimize $\ab$ jointly with support vectors, which is faster than optimizing basis and solving coordinates alternatively.
The gradients can be computed very efficiently using matrix operations.
Since gradient descent on support vectors is equivalent to moving these support vectors in a continuous space, thereby generating $m$ \emph{new} support vectors, we refer to these newly generated support vectors as \emph{gradient support vectors}.
We denote this combined method of LARS-SVM and gradient support vectors as Compressed Vector Machine (CVM).
Because the optimization problem in~(\ref{eq:gradient}) is non-convex with respect to $\x_i$, %
we initialize our algorithm with the basis $\Kb_m$ and coordinates $\alpha_m$ returned in the LARS-SVM solution.

In practice, it may be desirable to optimize both the SVM cost parameter $C$ and any kernel parameters (e.g. $\sigma^2$ in the RBF kernel) for the final CVM model.
Additionally, it may be preferable to optimize CVM constrained by the validation accuracy of the compressed model rather than the size of the support vector budget.
Constrained Bayesian optimization \citep{gardner2014bayesian} supports efficient constrained joint hyperparameter optimizations of this type.
Additionally, the L1-penalized support vector selection in the LARS-SVM step may benefit from recent work on highly parallel Elastic Net solvers \citep{zhou2015reduction}.

\section{Results}
\label{sec:results}
\label{sec:cvm_results}
In this section, we first demonstrate Compressed Vector Machine (CVM) on a synthetic data set to graphically illustrate each step in the algorithm. We then evaluate CVM on several medium-scale real-world data sets.

\begin{figure}[t]
\centerline{
\includegraphics[width = 1\textwidth]{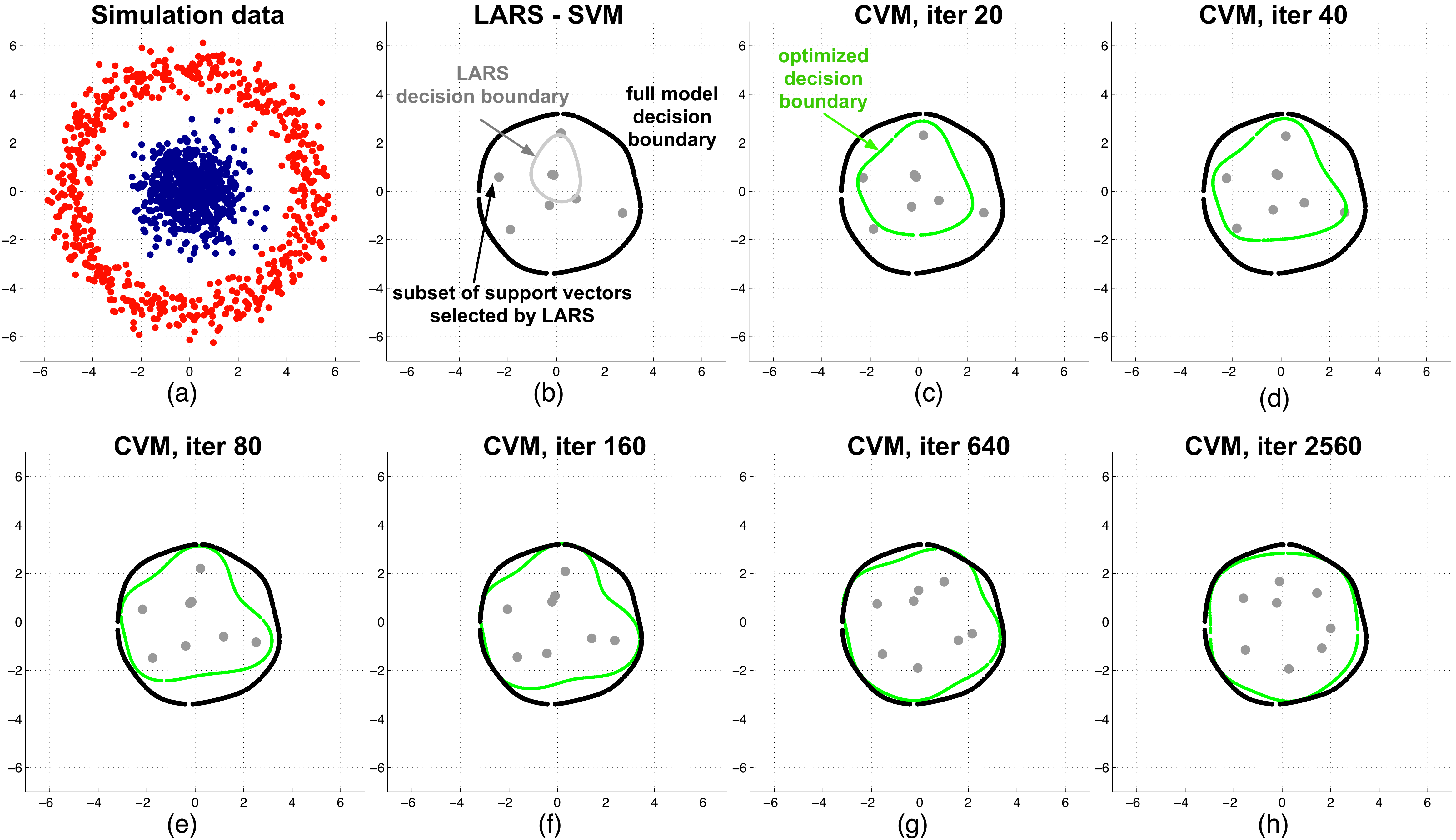}
}
\caption{Illustration of each step of CVM on a synthetic data set. (a) Simulation inputs from two classes (red and blue). By design, the two classes are not linear separable. (b) Decision boundary formed by full SVM solution (black curve). A small subset of support vectors picked by LARS (gray points) and the compressed decision boundary formed by this subset of support vectors (gray curve). (c-h) Optimization iterations. The gradient support vectors are moved by the iterative optimization. The optimized decision boundary formed by gradient support vectors (green curve) gradually approaches the one formed by full SVM solution.}
\label{fig:simul}
\end{figure}

\textbf{Synthetic data set.} The data set contains 600 sample inputs from two classes (red and blue), where each input contains two features. The blue inputs are sampled from a Gaussian distribution with mean at the origin and variance $1$, and red inputs are sampled from a noisy circle surrounding the blue inputs. As shown in Figure \ref{fig:simul}(a), by design the data set is not linearly separable. For simplicity, we treat all inputs as training inputs. To evaluate CVM, we first learn an SVM with the RBF kernel from the full training set. 
We plot the resulting optimal decision boundary in Figure \ref{fig:simul}(b) with a black curve.
In total, the full model has $80$ support vectors.

To compress the model, we first select a subset of support vectors by solving LARS-SVM optimization~(\ref{eq:readyforlars}). Specifically, we compress the model to $10\%$ of its original size, $8$ support vectors, by running LARS for $8$ iterations. The $8$ LARS-SVM support vectors are shown in Figure~\ref{fig:simul}(b) as solid gray points, and the approximate LARS-SVM decision boundary is shown by the gray curve.

Since the subspace formed by $8$ support vectors is heavily restricted by the discrete training input space, the approximation is poor. To overcome this problem, we search for a better subspace or basis in a continuous space, and perform gradient descent on support vectors by optimizing~(\ref{eq:gradient}). In Figure \ref{fig:simul}(c-h), we illustrate the optimization with updated support vector locations and optimized decision boundaries as we gradually increase the number of iterations. The resulting \emph{gradient support vectors} are shown as gray points and the new optimized decision boundaries formed from these new gradient support vectors are shown by green curves. After $2560$ iterations, as shown in Figure \ref{fig:simul}(h), we can observe that the optimized decision boundary (green) is very close to the boundary captured in the full model (black). These optimized decision boundaries demonstrate that moving a small subset of support vectors in a continuous space can efficiently approximate the optimal decision boundary formed by full SVM solution, supporting effective SVM model compression.

\begin{table}[h!!!]
    \tabcolsep 5pt
    \small
	\center	
\center
\begin{tabular}{|l|c|c|c|c|c|c|c|c|c|}
\hline {\bf Statistics} & {\bf Pageblocks} & {\bf Magic} & {\bf Letters} & {\bf 20news} & {\bf MNIST} & {\bf DMOZ} \\
\hline
\hline \#training exam. & 4379 & 15216 & 16000 & 11269 & 60000 & 7184 \\
\hline \#testing  exam. & 1094 & 3804 & 4000 & 7505 & 10000 & 1796  \\
\hline \#features & 10 & 10 & 16 & 200 & 784 & 16498 \\
\hline \#classes  & 2 & 2 & 26 & 20 & 10 & 16 \\
\hline
\end{tabular}
\caption{Statistics of all six data sets.}
\label{table:stats}
\end{table}

\begin{figure}[t]
\centerline{
\includegraphics[width = .705\textwidth]{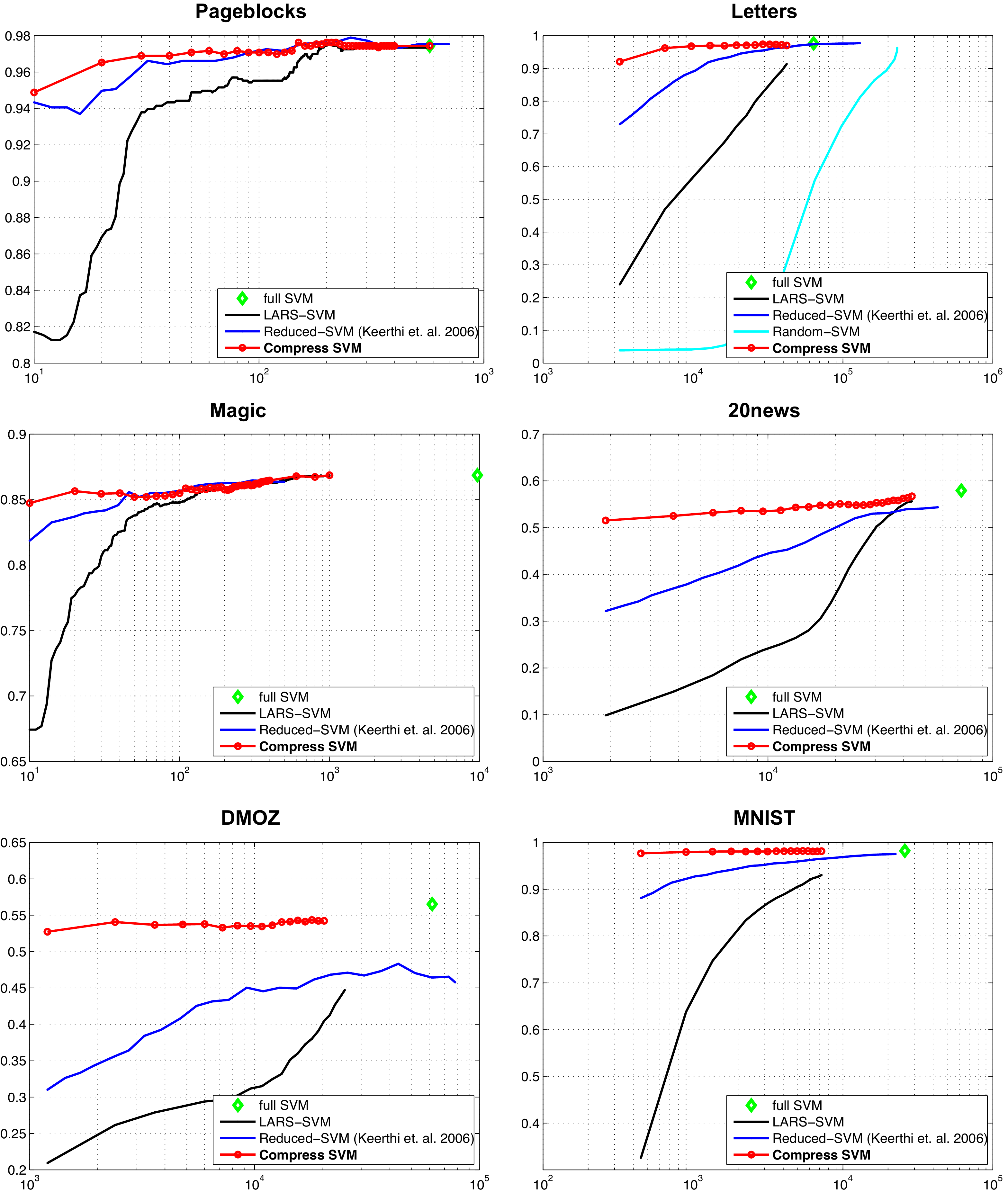}
}
\caption{Accuracy versus number of support vectors (in log scale).}
\label{fig:allsets}
\end{figure}

\textbf{Large real-world data sets.} To evaluate the performance of CVM on real-world applications, we evaluate our algorithm on six data sets of varying size, dimensionality and complexity. Table~\ref{table:stats} details the statistics of all six data sets.
We use LibSVM \cite{chang2011libsvm} to train a regular RBF kernel SVM using regularization parameter $C$ and RBF kernel width $\sigma$ selected on a $20\%$ validation split.
For multi-class data sets, we use the one-vs-one multi-class scheme.
The classification accuracy of test predictions from this SVM model serves as a baseline in Figure~\ref{fig:allsets}(full SVM).

Given the full SVM solution, we run CVM in two steps. First, we use LARS solve the optimization problem in~(\ref{eq:readyforlars}) using all support vectors from the original SVM model. An initial compressed support vector set is selected with a target compressed size (\emph{e.g.} $10$ out of $500$ support vectors). The selected support vectors serve as the second baseline in Figure~\ref{fig:allsets}(LARS-SVM). Second, we shift these support vectors in a continuous space by optimizing~(\ref{eq:gradient}) w.r.t. the input support vectors and the corresponding Lagrange multipliers $\ab$, generating gradient support vectors. This final set of gradient support vectors constitutes the CVM model. To show the trend of accuracy/cost performance, we plot the classification accuracy for CVM after adding every $10$ support vectors. Figure~\ref{fig:allsets} shows the performance of CVM and the baselines on all six data sets.

\textbf{Comparison with prior work.}
Figure~\ref{fig:allsets} also shows a comparison of CVM with \emph{Reduced-SVM}~\cite{keerthi2006building}. This algorithm takes an iterative two phase approach. First a set of support vectors is heuristically selected from random samples of the training set and added to the existing set of support vectors (initially empty). Then, the model weights are optimized by an SVM with the quadratic hinge loss. The algorithm alternates these two steps until the target number of support vectors is reached.

As shown in the Figure~\ref{fig:allsets}, CVM significantly improves over all baselines. Compared to the current state-of-the-art, Reduced-SVM, CVM has the capability of moving support vectors, generating a new basis, and learning a highly approximated basis to match the decision boundaries formed by the full SVM solution. It is this ability that distinguishes CVM from other algorithms when the evaluation budget is low. Across all data sets, CVM maintains close to the same accuracy as the full SVM with merely $10\%$ of the support vectors.

\section{Acknowledgments}
\label{sec:ack}
Most of this work was previously invented by Burges and Sch\"olkopf~\cite{burges1997improving} whose research was truly visionary at the time. 

\bibliography{compsvm}
\bibliographystyle{plain}
\end{document}